%% file: main_nips.tex
\documentclass{article}

\usepackage[preprint]{nips_2018}

\usepackage[utf8]{inputenc} 
\usepackage[T1]{fontenc}    
\usepackage{hyperref}       
\usepackage{url}            
\usepackage{booktabs}       
\usepackage{amsfonts}       
\usepackage{nicefrac}       
\usepackage{microtype}      
\usepackage{graphicx}

\usepackage{graphicx}
\usepackage{amsmath}
\usepackage{mathrsfs}
\usepackage{appendix}

\usepackage{array}
\usepackage{multirow}
\usepackage{tabu}

\font\fiverm=cmr5
\def\R{\mathbf{R}}
\def\N{\mathbf{N}}
\def\(#1){[\hbox{$\mkern1mu\thickmuskip=\thinmuskip#1\mkern1mu$}]} 
\def\ast{\mathop{\hbox{\lower 1.5pt\hbox{$\buildrel x\over *$}}}}

\def\R{{\bf R}}
\def\N{{\bf N}}
\def\feat{C\mskip -10mu\lower-2pt\hbox{\fiverm 1}\,}
\def\Feat#1{C\mskip -10mu\lower-2pt\hbox{\fiverm #1}\,}
\def\dts{\mathinner{\ldotp\ldotp}} 
\def\dash---{\thinspace---\hskip.16667em\relax} 
\let\del=\partial
\usepackage{epsfig}

\DeclareGraphicsExtensions{.eps,.pdf,.jpg,.gif}

\setcitestyle{numbers}

\title{Motion Invariance in Visual Environments}

\author{
  Alessandro Betti\\
  University of Florence\\
  Florence, Italy\\
  \texttt{alessandro.betti@unifi.it}\\
  \And
  Marco Gori\\
  SAILab\\
  University of Siena\\
  Siena, Italy\\
  \texttt{marco@diism.unisi.it}\\
  \And
  Stefano Melacci\\
  SAILab\\
  University of Siena\\
  Siena, Italy\\
  \texttt{mela@diism.unisi.it}\\
}

\begin{document}

\maketitle

\begin{abstract}
The puzzle of computer vision might find new challenging solutions when we realize that most successful methods are working at image level, which is remarkably more difficult than processing directly visual streams, just as happens in nature. In this paper, we claim that their processing naturally leads to formulate the motion invariance principle, which enables the construction of a new theory of visual learning based on convolutional features. The theory addresses a number of intriguing questions that arise in natural vision, and offers a well-posed computational scheme for the discovery of convolutional filters over the retina. They are driven by the Euler-Lagrange differential equations derived from the principle of least cognitive action, that parallels laws of mechanics. Unlike traditional convolutional networks, which need massive supervision, the proposed theory offers a truly new scenario in which feature learning takes place by unsupervised processing of video signals. 
An experimental report of the theory is presented where we show 
that features extracted under motion invariance 
yield an improvement 
that can be assessed by measuring information-based indexes. 
\end{abstract}

\section{Introduction}
\input Introduction

\section{The principle of least cognitive action}
\input CN-architecture
\section{Discretization on the retina and Euler-Lagrange equations}
\input CAL
\section{Experiments}
\input exp.tex

\section{Conclusions}
\input Conclusions

\bibliography{nn,corr}
\bibliographystyle{unsrt}
\end{document}

%% file: Introduction.tex
While the emphasis on a general theory of vision was already 
the main objective at the dawn of the discipline~\cite{Marr82}, computer vision has evolved without a systematic exploration of foundations in the framework of machine learning. In particular, in most cases, computer vision is regarded just as an application of machine learning
. 
When the target is moved to unrestricted visual environments and the emphasis is shifted from huge labelled databases to a human-like protocol of interaction, we need to go beyond the current peaceful interlude that we are experimenting in vision and machine learning. 
So far, the semantic labeling of pixels of a given video stream has been mostly carried out at frame level. This seems to be the natural outcome of well-established pattern recognition methods working on images, which have given rise to nowadays emphasis on collecting big labelled image databases (e.g.~\cite{imagenet_cvpr09}) 
with the purpose of devising and testing challenging machine learning algorithms. While this framework is the one in which most of the state of the art object recognition approaches have been developing, we argue that there are strong arguments to start exploring the more natural visual interaction that animals experiment in their own environment. 
This leads to process videos instead of image collections 
that is very much related to the growing interest of {\em learning in the wild} that has been explored in the
last few years (see. e.g.~\url{https://sites.google.com/site/wildml2017icml/}).

A crucial problem that has been recognized by Poggio and Anselmi~\cite{Poggio:2016:VCD} is the need to incorporate visual invariances into deep nets that go beyond simple translation invariance 
that is currently characterizing
convolutional networks. They propose an elegant mathematical framework 
on visual invariance and enlighten some intriguing neurobiological connections. 
Overall, the ambition of extracting distinctive features from vision poses a challenging task. 
While we are typically concerned with feature extraction methods that are independent of classic geometric transformation,
it looks like we are still missing the fantastic human skill of capturing distinctive features to recognize ``ironed and rumpled shirts'', for example. There is no apparent difficulty to recognize shirts by keeping the recognition coherence in case we roll up the sleeves, or we simply curl them up into a ball for the laundry basket. Of course, there are neither rigid transformations, like translations and rotation, nor scale maps, that transforms an ironed shirt into the same shirt thrown into the laundry basket. 
In this paper, we claim that motion invariance can in fact capture all we need.
Translation and scale invariance, that have been the subject of many studies \cite{LoweCV2004,DBLP:journals/cviu/GoriLMM16}, are in 
fact examples of invariances that can be fully gained whenever we develop the 
ability to detect features that are invariant under motion. 
For instance, the moving of the finger experimented by infants leads them to 
enforce a natural invariance: it will  become bigger and bigger as it approaches their 
face, but it is still their inch, which requires to impose a consistent decision. 
Clearly, translation, rotation, and complex deformation invariances  derive from
motion invariance. Humans life always experiments motion, so as the gained visual
invariances naturally arise from motion invariance. Animals with foveal eyes also
move quickly the focus of attention when looking at fixed objects, which means that they
continually experiment motion. Hence, also in case of fixed images, 
conjugate, vergence, saccadic, smooth pursuit, and vestibulo-ocular movements lead to acquire visual information from relative motion.  We claim that the 
production of such a continuous visual stream naturally drives
feature extraction, since the corresponding convolutional features are expected not to change during motion. 
The enforcement of this consistency condition creates a mine of visual data
during animal life. 
Of course, we need to compute the optical flow at pixel level so as to enforce
the consistency of all the extracted features. Early studies 
on this problem~\cite{HornAI1981}, along with recent related 
improvements (see e.g.~\cite{Baker:2011})
suggest to start computing the velocity field by enforcing brightness invariance. 
As the optical flow is gained, it is used to enforce motion consistency on the
visual features. Interestingly, the theory we propose is quite related to the
variational approach that is used to determine the optical flow in~\cite{HornAI1981},
but the joint feature development of the features can also be used to reinforce 
motion estimation. 
It is worth mentioning that an effective visual system must also develop 
features that do not follow motion invariance. These kind of features can be 
conveniently combined with those that are discussed in this paper with the
purpose of carrying out high level visual tasks.


The visual features are  derived in the framework of the principle of 
cognitive action~\cite{DBLP:journals/tcs/BettiG16}, which gives rise  to a
time-variant differential equation, where the Lagrangian coordinates
correspond with the values of the convolutional filters. 
The learning process can be interpreted in the framework of the minimization
of the cognitive action that offers a self-consistent framework. 

%% file: CN-architecture.tex
We consider the mechanisms that give rise
to the construction of local features for any pixel $x \in X$ of the retina, at any
time $t$. These features, along with the video itself, can be regarded as 
visual fields, that are defined on the retina and on a given  horizon of
time  $[0\dts T]$.
A set of symbols are extracted
at every layer of a deep architecture, so as each pixel\dash---along with its context\dash---
turns out to be represented by the list of symbols extracted at each layer.
The computational process that we define involves the video
as well as appropriate vector fields 
that 
are defined on the domain $D=X\times [0\dts T]$.
In what follows, points on the retina will be represented
with two dimensional vectors $x=(x_1,x_2)$ on a defined 
coordinate system
.
The temporal coordinate is usually denoted by
$t$, and, therefore, the video signal on the pair $(x,t)$ is $C(x,t)$. 
The color field can be thought of as a special field that is characterized by $m$
 components for each single pixel ($m=3$ for RGB) . 
We are concerned with the problem of extracting visual features that, unlike
the components of the video, express the information associated with 
the pair $(x,t)$ and its spatial context. Basically, one would like to extract
visual features that characterize the information in the neighborhood of pixel $x$.
A possible way of constructing this kind of features is to define\footnote{Throughout the
paper we use the Einstein convention to simplify the equations.}
\begin{equation}
	\feat_i(x,t)=\frac{1}{n}+
	\sum_{j=0}^{m-1}\int_X dy \ \varphi_{ij}(x,y,t)C_j(y,t)
\label{Kernel-contex-def}
\end{equation}
Here we assume that $n$ symbols are generated from the $m$
components of the video. 
Notice that the kernel $\varphi(x,y,t)$ is responsible of expressing the spatial dependencies.
It is worth mentioning that 
whenever $\varphi(x,y,t) \leadsto \varphi(x-y,t)$ the above definition
reduces to an ordinary spatial convolution. The computation 
of   $\feat_i(x,t)$ yields a field with $n$ features
, and Eq.~(\ref{Kernel-contex-def}) can be used for carrying out a piping scheme
where a new set of features $\Feat2$ is computed from $\Feat1$. Of course,
this process can be continued according to a deep computational structure 
with a homogeneous convolutional-based computation, 
which yields the features $C\mskip -10mu\lower-3pt\hbox{\fiverm z}\,$
at the $z$-th convolutional layer. 
The theory proposed in this paper focuses on the construction of any of these
convolutional layers which are expected to provide higher and higher abstraction
as we increase the number of layers. 
The {\it filters\/}  $\varphi$ are what completely determines the
features $\feat_i(x,t)$. In this paper we formulate a theory for the discovery of $\varphi$
that is based on three driving principles, that are described below.

\paragraph{Optimization of information-based indices}
	Beginning from the color field $C$, we attach
	a symbol $y_i \in \Sigma$ of a discrete vocabulary
	to pixel $(x,t)$ with probability $\feat_i(x,t)$. 
	This is obtained by estimating the random variable $F(X)$
	where $F$ is the map that the agent is expected to learn, that is
	defined on the basis of Eq.~\ref{Kernel-contex-def}.
	The conditional entropy $S(Y\mid X,T,F)$ is given by 
	$S(Y\mid X,T,F)=-\int_\Omega \sum_{i=1}^n dP_{X,T,F} \  p_i
	\log p_i\, $
where $p_i$ is the conditional probability of $Y$ conditioned
to the values of $X$, $T$ and $F$, $dP_{X,T,F}$ is the joint measure of
the variable $X,T,F$, and $\Omega$ is a Borel set in the $(X,T,F)$
space. 
We assume that $\forall x,t: \ \feat_i(x,t)$ is subject to the probabilistic constraints
$\sum_i \feat_i(x,t)=1$ (normalization) and 
$0\le\feat_i(x,t)\le1$ (positivity). 
We can rewrite the conditional entropy as 
	$S(Y\mid X,T,F)=-\int_Dd\mu(x,t) \ \sum_{i=1}^n\feat_i(x,t)\log
	\feat_i(x,t)$,
        where $\mu(x,t)$ is a space-time measure.
Clearly, we want to keep the conditional entropy as small as possible
so as to develop dominating features. At the same time we must ensure that 
the entropy of  variable $Y$,
\begin{equation}
	S(Y)=-\sum_{i=1}^n \Pr(Y=y_i)\log\Pr(Y=y_i) \ ,
\label{BalEntropybyPr}
\end{equation}
must be as high as possible, since this ensures the development of 
all the features associated with the alphabet of symbols.
If we use the law of total probability to express $\Pr(Y=y_i)$ 
in terms of the conditional probability $p_i$ and  use
the above assumptions we get\\
	$\Pr(Y=y_i)=\int_\Omega dP_{X,T,F} \ p_i=\int_D d\mu(x,t) \, \feat_i(x,t)$.
Then
\begin{equation}
	S(Y)=-\sum_{i=1}^n \Big(\int_D d\mu(x,t) \, \feat_i(x,t)\Big)
	\log\Big(\int_D d\mu(x,t) \, \feat_i(x,t)\Big).
\label{BalancingEntropybyC}
\end{equation}
To sum up (see the Supplementary Material for further details on the computation of $S(Y)$), the index ${\cal I}(\varphi)=S(Y)-S(Y\mid X,T,F)$, which is somewhat 
related to the classic Shannon mutual information, must be maximized \cite{gori2012information,DBLP:journals/tnn/MelacciG12}. 
	
\paragraph{Motion invariance}
If we focus attention
on a the pixel $x$ at time $t$, which moves according to the trajectory $x(t)$
then  $\feat(x(t),t) = c$, being $c$ a constant. 
This ``adiabatic'' condition is thus expressed by the condition
$d\feat/dt=0$, which yields 
\begin{equation} 
 	\partial_t \feat_i+v_j\partial_j \feat_i=0,
\label{LocalMIEq}
\end{equation}
where $v\colon D\to\R^2$ is the velocity field that we assume to be given,
and $\partial_k$ is the partial derivative with respect to $x_k$.
When replacing $\feat_i$ as stated by Eq.~(\ref{Kernel-contex-def}) we get
\begin{align*}
	{\cal M}(\varphi)= \int_Xdy\,\big(\partial_t\varphi_{ij} C_j+\varphi_{ij}
	\partial_t C_j +  v_k \partial_k \varphi_{ij} C_j\big)=0,
\end{align*}
which holds for any $(t,x) \in D$.
Notice that this constraint is linear in the field $\varphi$.
This can be interpreted by stating that learning under
motion invariance consists of determining elements of the
kernel of the function 
${\cal M}(\varphi)$. 
Clearly, the learning process is expected to keep the value of
${\cal M}(\varphi)$ as small as possible.

\paragraph{Parsimony principle}
	Like any principled formulation of learning, 
	we require the filters to obey the parsimony principle. 
	Amongst the philosophical implications, it also favors the development of a unique 	
	solution.
	Given the filters $\varphi$, 
	there are two parsimony terms, one 
	${\cal P}(\varphi)$, that penalizes abrupt spatial
	changes, and another one, ${\cal K}(\varphi)$  that penalizes 
	quick temporal transitions.
	Ordinary regularization issues suggest to discover functions $\varphi_{ij}$ such that
\[
  \lambda_P{\cal P}+\lambda_K{\cal K}=\frac{\lambda_{P}}{2}
  \int_D dtdx\, h(t) (P_{x}
  \varphi_{ij}(x,t))^{2}
  +\frac{\lambda_{K}}{2} \int_{D} dtdx\, h(t)
(P_{t} \varphi_{ij}(x,t))^{2},\]
is ``small'', where $P_{x}, P_{t}$ are spatial and temporal differential operators, and  $\lambda_{P},  \lambda_{K}$ are non-negative reals. We assumed an ergodic translation of
$d\mu$, that, in this case, only involves the temporal factor
$h(t)$.

Overall, the process of learning is regarded as the minimization of the {\em cognitive action}
\begin{equation}
	{\cal A}(\varphi)= - {\cal I}(\varphi) + \lambda_{M} {\cal M}(\varphi)
	+\lambda_{P} {\cal P}(\varphi)
	+ \lambda_{K} {\cal K}(\varphi),
\label{CognitiveActionEq}
\end{equation}
where $\lambda_{M},\lambda_{P},\lambda_{K}$ are positive multipliers.
While the first and third principles are typically adopted in classic unsupervised learning, 
motion invariance does characterize the approach followed in this paper.
Of course, there are visual features that do not obey the motion invariance principle. 
Animals easily estimate the distance to the objects in the environment, a property that clearly
indicates the need for features whose value do depend on motion. The perception of
vertical visual cues, as well as a reasonable estimate of the angle with respect to the
vertical line also suggests the need for features that are motion dependent. 
Basically, the process of learning consists of solving the variational problem 
$\hat{\varphi} = \arg \min_{\varphi} {\cal A}(\varphi)$
(see the Supplementary Material for details). 
As it will be shown in the following, in our multi-layer implementation the minimization of ${\cal A}(\cdot)$ takes place at each layer of the architecture, involving the filters of the considered layer only, relying on a piping scheme that is inspired to developmental learning issues. 

%% file: CAL.tex
\label{discretization-section}
The field theory of the previous section can be approximated
over the discrete (and bounded) retina $X^\sharp$, where the video frames are represented.
Instead of the fields $\varphi_{ij}(x,t)$, we have a bunch of functions of
time
%
$\varphi_{ij x}(t)$, indexed by the point on the retina $x$ other than the
filter/feature index $i$ and the input channel index $j$. Similarly, the color field will
be replaced by $C_{kx}(t)$.
Using Einstein notation  we have that the discretized form
of the feature fields is $\feat_{ix}^\sharp(t)=1/n+
\varphi_{iky}(t)C_{k(x-y)}(t)$, where the sum in $y$
is performed over $X^\sharp$. The two pieces of the motion invariance term (\ref{LocalMIEq}) become
\[\begin{split}
    &\del_t\feat_i(x,t)\mathrel{\mathop{\longrightarrow}^{\rm disc\,\,\,}}
    \dot\feat_{ix}^\sharp(t)=\dot \varphi_{iky}(t)C_{k(x-y)}(t)+
    \varphi_{iky}(t)\dot C_{k(x-y)}(t);\cr
    &\del_j\feat_i(x,t)\mathrel{\mathop{\longrightarrow}^{\rm disc\,\,\,}}
    \Delta_j \feat_{ix}^\sharp(t)= \varphi_{iky}(t)\Delta_jC_{k(x-y)}(t),
\end{split}\]
where $\Delta$ the spatial gradient operator.
Such term of motion invariance becomes a
quadratic form in $\varphi$ and $\dot \varphi$.
The other relevant terms of the theory (entropy,
relative entropy) are trivially functions of $\varphi(t)$ and $t$.
We assume filters to have a unique and finite size for all the features. As a consequence, for each feature $i$, we can flatten the filter $\varphi_{ij x}$ into a vector, and concatenate the $n$ filter-vectors into $q$.
We selected a second order term to implement the parsimony principle, $h \cdot (\frac{\alpha}{2}|\ddot q|^2+  \frac{\beta}{2}|\dot q|^2+\gamma\ddot q\cdot \dot q+\frac{k}{2}|q|^2)$, being $\alpha,\beta,\gamma,k$ positive constants.
If we make the entropy term local in time, and evaluate the first variation of the discretized cognitive action, the differential Euler-Lagrange (EL) equations are (for the sake of simplicity, we skip the derivations, see Supplementary Material for all the details):
\begin{equation}\begin{split}
\hat \alpha q^{(4)}+2\dot{\hat\alpha}q^{(3)}+(\ddot{\hat\alpha}+\dot{\hat\gamma}-\hat R)
\ddot q &-\left(\dot{\hat R}-\ddot{\hat\gamma}-\lambda_M\hat N^\natural+\lambda_M(\hat N^\natural)'\right)\dot q\cr
&-\left((\dot{\hat N}^\natural)'-\hat Z\right)q +\frac{1-\lambda_C}{n}b+\nabla_q w(t,q)=0,
\cr\end{split}
\label{Gen-q-DiffEq}
\end{equation}
where $q^{(4)},q^{(3)}$ are the fourth and third derivatives of $q$ over time, $\lambda_C$ is a positive constant, and we have used the notation $\hat f=h f$ (so that for example $\ddot{\hat \alpha}=\ddot h \alpha$). In order to define the other terms, we introduce the notation $\Gamma_x$ to indicate the area (volume if $m>1$) of the input signal centred around $x$, of the same size of the filters, flattened into a vector. We have $R:=\beta+\lambda_M M^\natural$, and the notation $A^\natural$ indicates a block-diagonal matrix whose blocks are $A$. The matrix $M$ is composed of $M_{al}:=\sum_{x\in X^\sharp} g_x \Gamma_x(a) \Gamma_x(l)$, and $g_x$ is a distribution over the retina. Analogously, we define $N_{al}:=\sum_{x\in X^\sharp} g_x r(a) \Gamma_x(l)$, being $r:=\dot{ \Gamma}_x + v \cdot \Delta \Gamma_x$. We have $Z:=k+B^\natural-\lambda_CM^\natural+\lambda_1 \tilde{M}+\lambda_M O^\natural$, where $\tilde{M}$ is a squared matrix composed of $m \times m$ repetitions of $M$, $\lambda_1$ is a positive constant, $B_{al}=b_a b_l$, and $b_a=\sum_{x\in X^\sharp} g_x \Gamma_x(a)$. The matrix $O$ is composed of $O_{al}=\sum_{x\in X^\sharp} g_x r(a) r(l)$. Finally, $ w(t,q):=\sum_{x\in X^\sharp} g_x  (\frac{1}{n} + \xi(q,  \Gamma_x))'  [\frac{1}{n} + \xi(q,  \Gamma_x)<0]$, where $\xi(q,  \Gamma_x)$ returns the $n$-length vector with the result of the convolutions of the $n$ filters with the input, $[\cdot]=1$  if the condition in brackets is true, otherwise it is 0, and it operates element-wise when a vector of conditions is provided.


In deriving equations some conditions arises naturally at $t=T$ (see the Supplementary Material for more details):
\begin{align}
\begin{split}
\label{BoundCondq0}
  \hat\alpha\ddot q(T)+\hat\gamma\dot q(T)=0,\quad
-\hat\alpha q^{(3)}(T)-\dot{\hat\alpha}\ddot q(T)+(\hat\beta+\lambda_M\hat M^\natural -\dot{\hat\gamma})
    \dot q(T)+\lambda_M(\hat N^\sharp)'q(T)=0
\end{split}
\end{align}

An interesting special case of these equations is that obtained with a null
signal $C\equiv0$.  With this assumption our equations (\ref{Gen-q-DiffEq}) become
\[\hat \alpha q^{(4)}+2\dot{\hat\alpha}q^{(3)}+(\ddot{\hat\alpha}+\dot{\hat\gamma}-\hat\beta)
\ddot q+(\ddot{\hat\gamma}-\dot{\hat\beta})\dot q+\hat kq=0.\] 
Now assume that $h(t)=e^{\theta t}$, with $\theta$ positive, then
\begin{equation}
	\alpha q^{(4)}+2\theta\alpha q^{(3)}+(\theta^2 \alpha+\theta\gamma-\beta)
  \ddot q+(\theta^2\gamma-\theta\beta)\dot q+kq=0.
\label{NightDynamics}
\end{equation}
In order to see whether this equation can be stable we need to apply the 
Routh-Hurvitz criterion.
For a fourth order ODE $q^{(4)}+aq^{(3)}+b\ddot q+c\dot q+dq=0$ this criterion reduces to check if
$a>0$, $b>0$, $0<c<ab$ and $0<d<(abc -c^2)/a^2$ that in our case means that
\begin{equation}
\alpha>0,\quad \beta>0,\quad \gamma>\frac{\beta}{\theta},\quad
0<k<\frac{(\beta-\gamma\theta)  [\beta-\theta(\gamma+2\alpha\theta)]}{4 \alpha}.
\label{StabilityConditions}
\end{equation}
So for example if we choose $\alpha=k=1/2$, $\gamma=2$ and
$\theta=\beta=1$ we obtain a stable equation. This being said it is
also crucial to notice that we have control over the important
parameter of the theory $\theta$ as long as you choose the
regularization parameters carefully. 

%% file: exp.tex



We implemented a solver for the differential equation of (\ref{Gen-q-DiffEq}) that is based on the Euler method with step size $\tau$. After having reduced the equation to the first order, the variables that were updated 
at each $t$ are $q$, $\dot q$, $\ddot q$, and $q^{(3)}$. The code and data we used to run the following experiments can be downloaded at \url{http://see.supplementary.material}, together with the full list of model parameters.
We randomly selected two real world video sequences from the Hollywood Dataset HOHA2 \cite{marszalek09}, that we will refer to as ``skater'' and ``car'', and a clip from the movie ``The Matrix'' (\textcopyright  Warner Bros. Pictures). The frame rate of all the videos is $\approx$ 25 fps (we set $\tau=1/25$), each frame was rescaled to $240\times110$ and converted to grayscale. Videos have different lengths, ranging from $\approx 10$ to $\approx 40$ seconds, and they were repeated in loop until $45,000$ frames were generated, thus covering a significantly longer time span. 
We randomly initialized $q(0)$, while the derivatives at time $t=0$ were set to $0$. We used the softmax function to force a probabilistic activation of the features, and computed the optical flow $v$ using an implementation from the OpenCV library. Convolutional filters cover squared areas of the input frame, and we set $g_x$ to be the uniform distribution.
All the results that we report are averaged over 10 different runs of the algorithms.

The video is presented gradually to the agent so as to favour the acquisition of small chunks of information. We start from a completely null signal (all pixel intensities are zero), and we slowly increase the level of detail and the pixel intensities, in function of $\phi(t) \in [0,1]$. In detail, 
$C(x,t) = \phi(t) [ \texttt{gauss}\left(1- \delta\phi(t)\right) \ast C^{o}(x,t) ]$, 
where $\ast$ is the spatial convolution operator, $C^{o}(x,t)$ is the source video signal, $\texttt{gauss}(\sigma^2)$ is a Gaussian filter of variance $\sigma^2$, and $\delta \in [0,1]$ is a customizable scaling factor. 
We start with $\phi(0)=0$, and then $\phi$ is progressively increased as time passes, $\phi(t+1) = \phi(t) + \eta(1-\phi(t))$ (we set $\eta = 0.0005$). 
We refer to the quantity $1-\phi$ as the ``blurring factor''.
In order to be able to (approximately) satisfy the conditions in Eq.~(\ref{BoundCondq0}) we need to keep the derivatives small, so we implement a
``reset plan'' according to which the video signal undergoes a reset
whenever the derivatives become too large. Formally, if $\|\dot q(t')\|^2\geq\epsilon_{1}$, or $\|\ddot q(t')\|^2\geq\epsilon_{2}$, or $\|q^{(3)}(t')\|^2\geq\epsilon_{3}$ then we forced $\phi(t')$ to $0$
($\epsilon_j=300 \cdot n$, for all $j$), and then we set to $0$ all the derivatives. 

\begin{figure}
\centering
\includegraphics[width=0.32\textwidth]{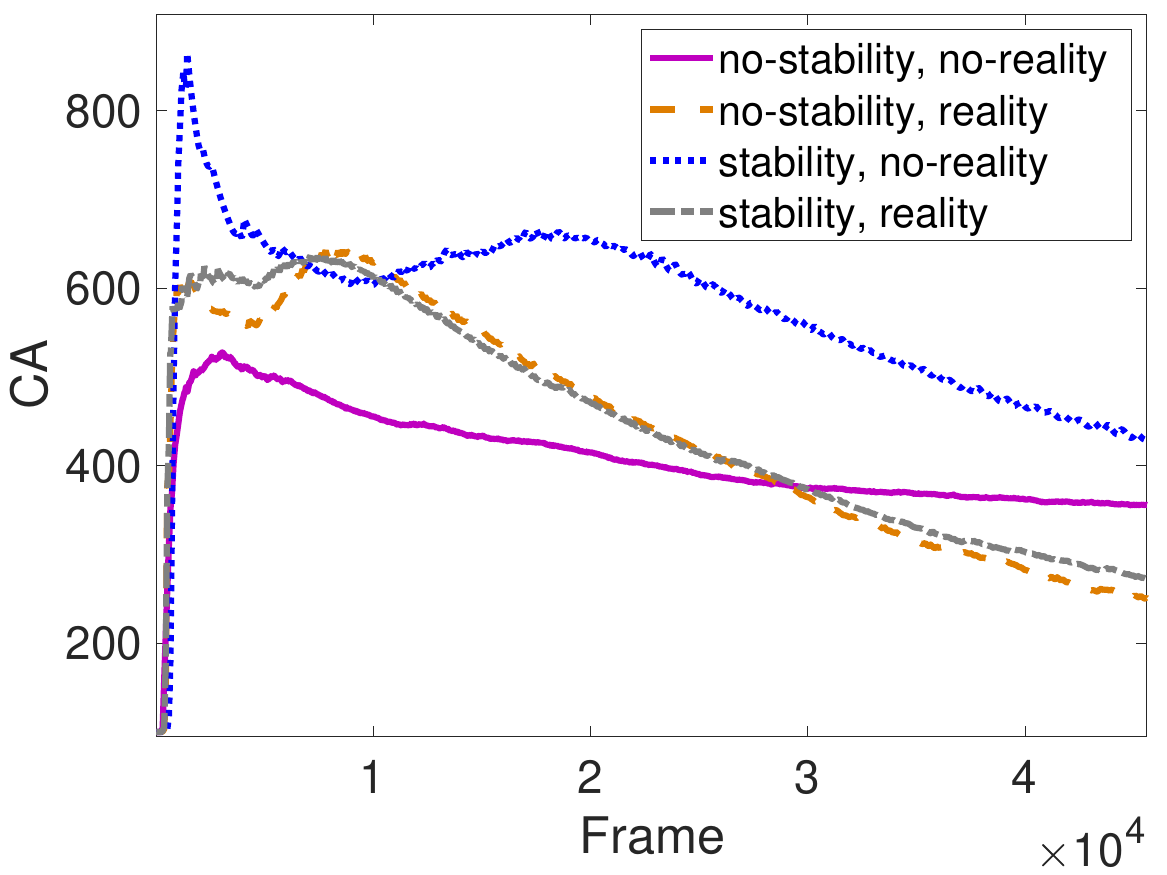}
\includegraphics[width=0.32\textwidth]{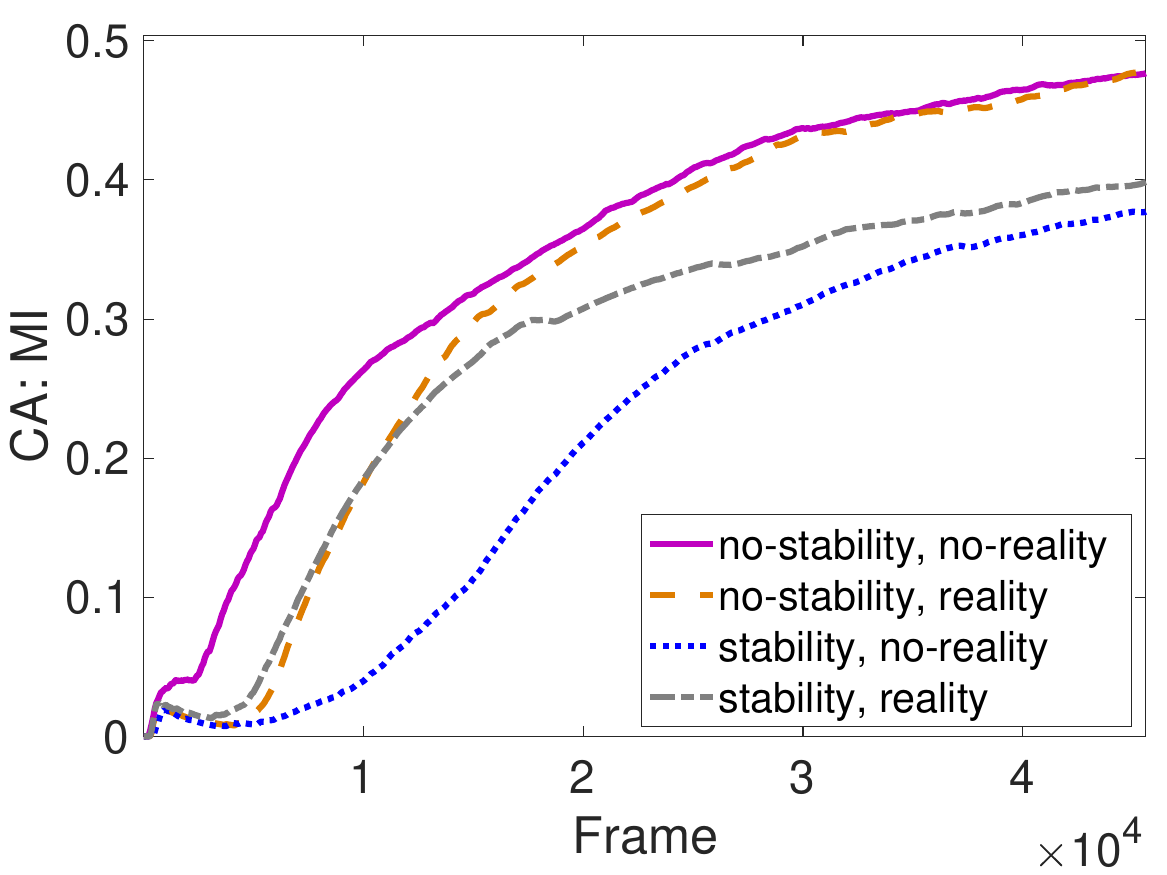}
\includegraphics[width=0.32\textwidth]{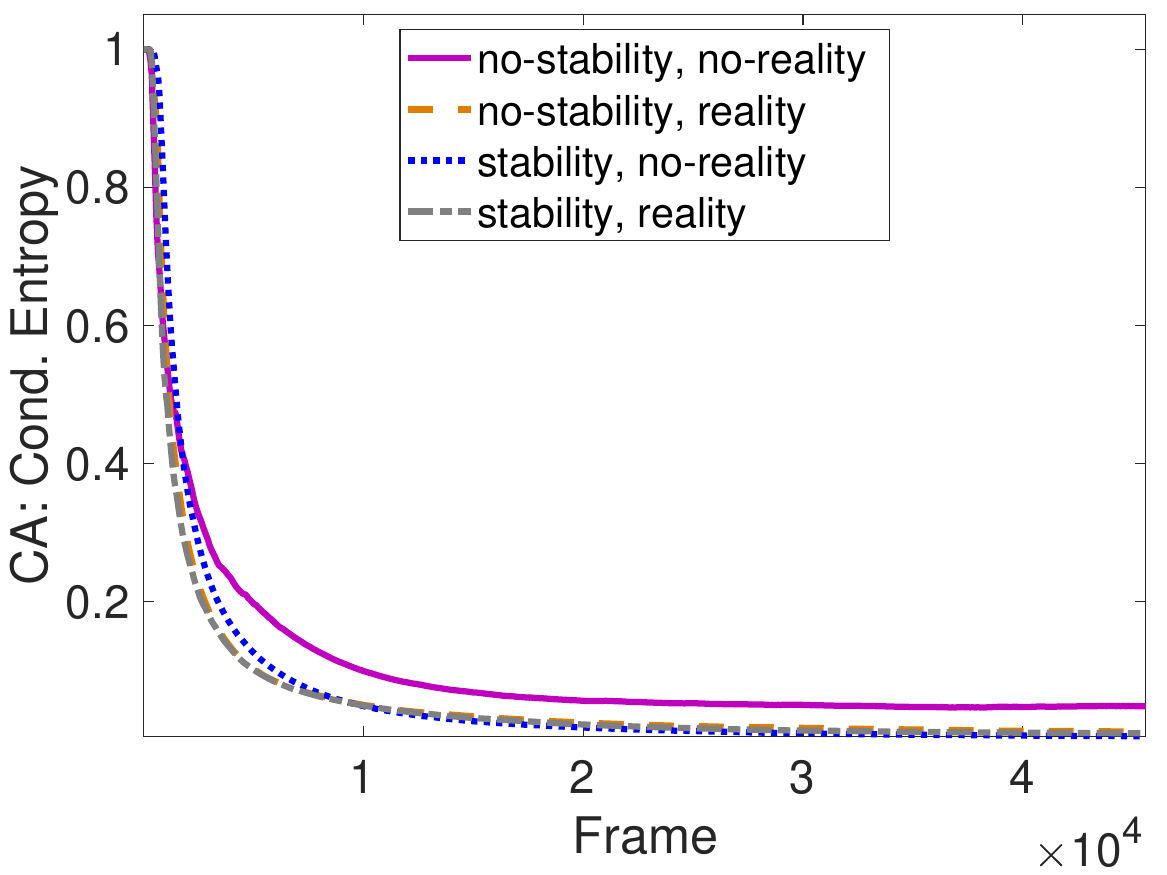}\\
\includegraphics[width=0.32\textwidth]{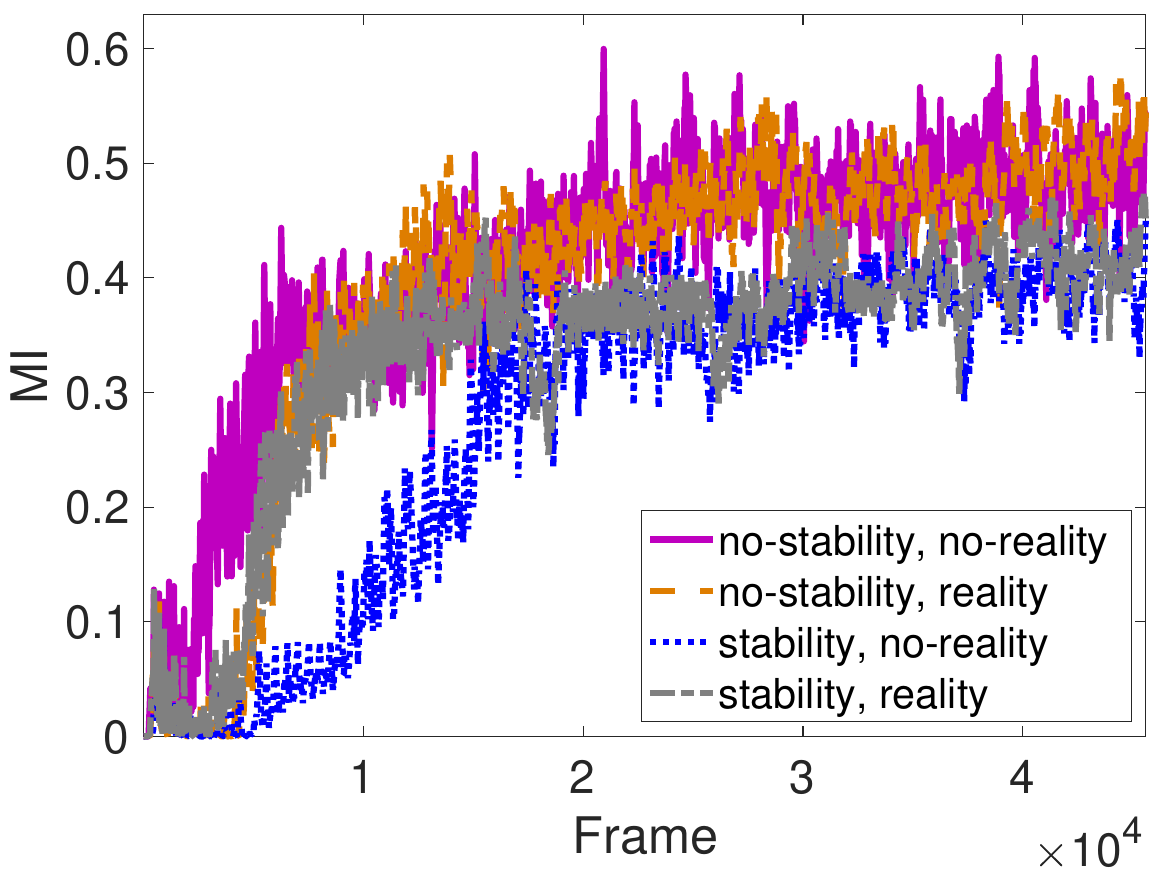}
\includegraphics[width=0.32\textwidth]{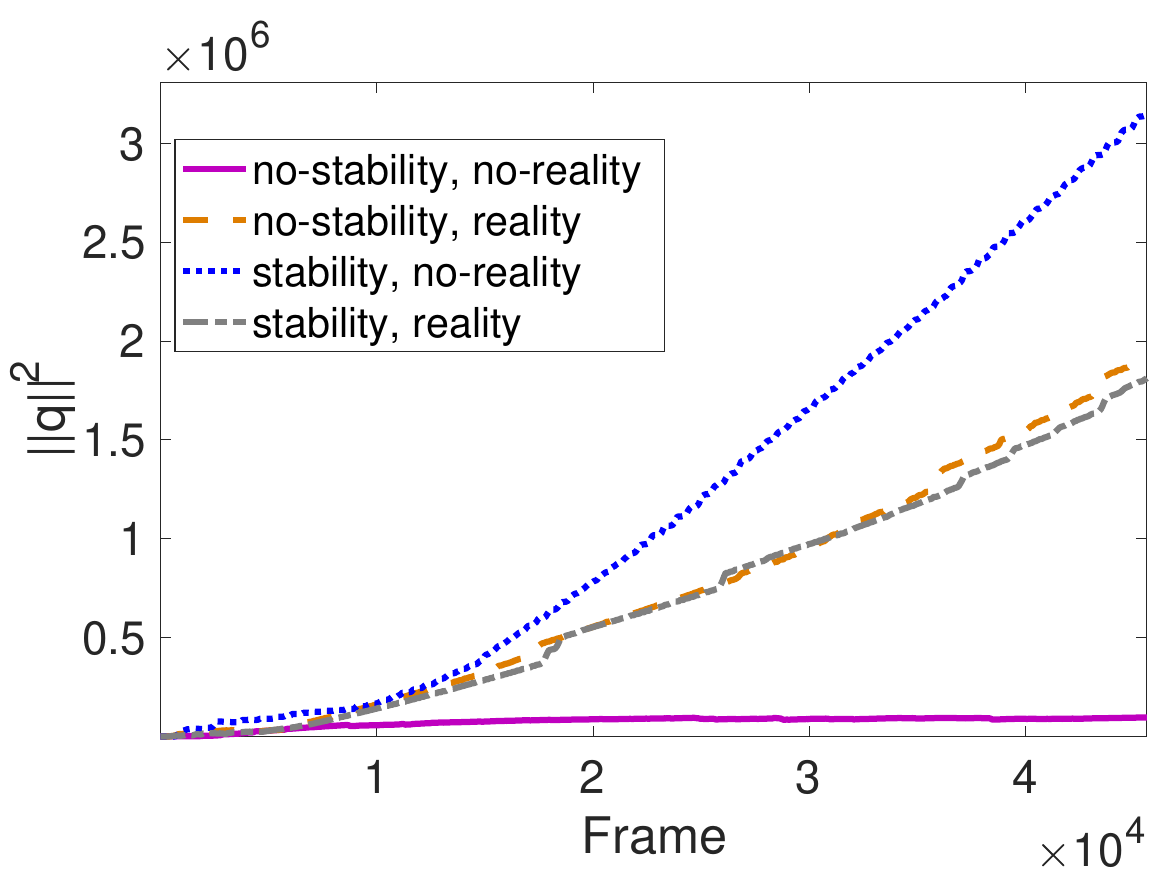}
\includegraphics[width=0.32\textwidth]{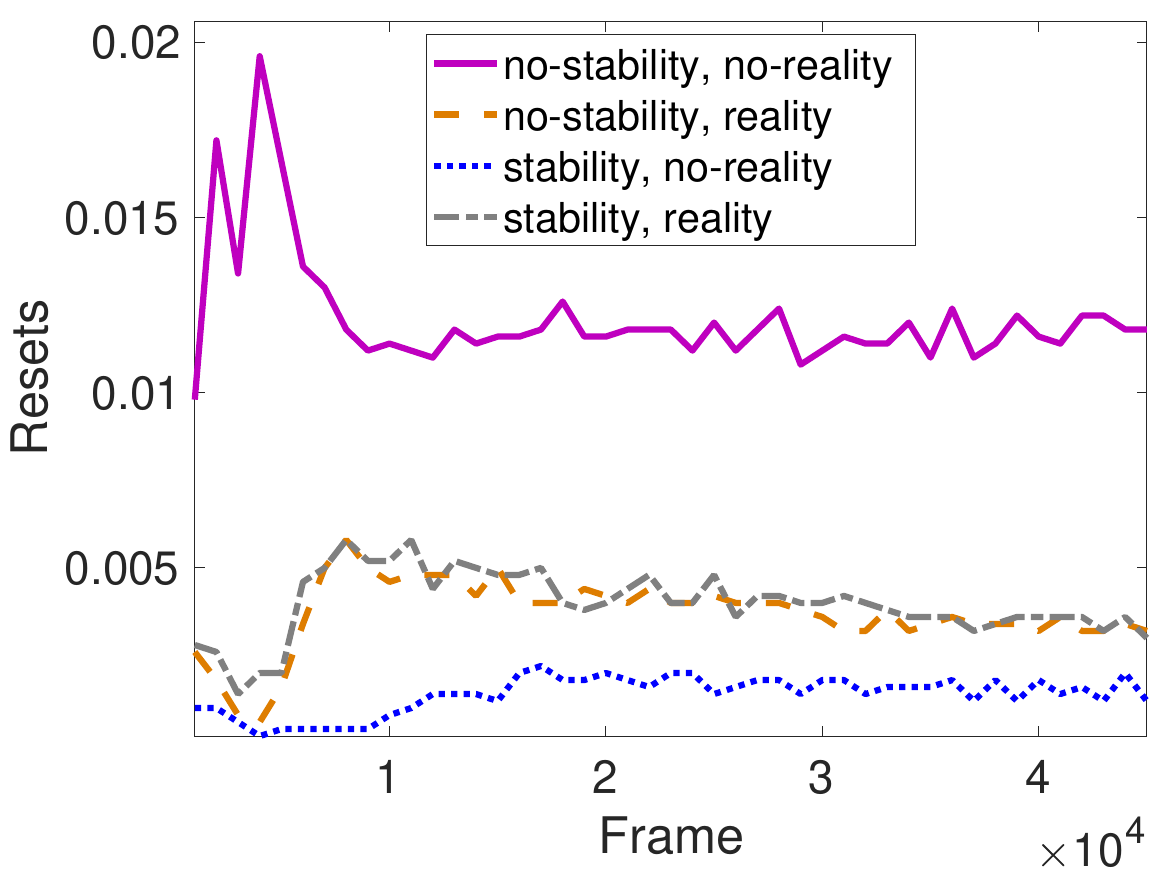}\\
\caption{Comparing 4 configurations of the parameters, characterized by different properties in terms of stability and reality of the roots of the characteristic polynomial. The input video is reproduced (in loop) for 45k frames (x-axis). From left-to-right, top-to-bottom we report the Cognitive Action (CA), the portion of the cognitive action that is about the Mutual Information (MI) (that we maximize), the portion that is about the Conditional Entropy, the MI per-frame, the norm of $q(t)$, and the fraction of ``reset'' operations performed every 1000 frames.}
\label{smallk} 
\end{figure}

Our experiments are designed (\textit{i}) to evaluate the dynamics of the cognitive action in function of different temporal regularities imposed to the model weights (parsimony), and then (\textit{ii}) to evaluate the effects of motion, that introduces a spatio-temporal regularization on single and multi-layer architectures.
When evaluating the temporal regularities, the cognitive action is composed by the entropy-based and parsimony terms only, and we experiment four instances of the set of parameters $\{\alpha, \beta, \gamma, k\}$. Each instance is characterized by the roots of the characteristic polynomial that lead to \textit{stable} or \textit{not-stable} configurations, and with only \textit{real} or also \textit{imaginary} parts, keeping the roots close to zero, and fulfilling the conditions of Eq.~(\ref{StabilityConditions}) when stability and reality are needed. These configurations are all based on values of $k \in [10^{-19}, 10^{-3}]$, while $\theta=10^{-4}$.
We performed experiments on the ``skater'' video clip, setting $n=5$  features, 
and chose filters of size $5\times5$. Results are reported in Fig.~\ref{smallk}. 
The plots indicate that there is an initial oscillation that is due to the effects of the blurring factor, that vanish 
after about 10k frames. The Mutual Information (MI) (${\cal I}$) portion of the cognitive action correctly increases over time
, and it is pushed toward larger values in the two extreme cases of ``no-stability, reality'' and ``no-stability, no-reality''. The latter shows more evident oscillations in the frame-by-frame MI value, due to the roots with imaginary part. In all the configurations the norm of $q$ increases over time (with different speeds), due to the small values of $k$, while the frequency of reset operations is larger in the ``no-stability, no-reality'' case, as expected. 

%
We evaluated the quality of the developed features by freezing the final $q$ of Fig.~\ref{smallk} and computing the MI index over a single repetition of the whole video clip, reporting the results in Tab.~\ref{nffsb} (a). This is the procedure we will follow in the rest of the paper when reporting numerical results in all the tables. We notice that, while in Fig.~\ref{smallk} we compute the MI on a frame-by-frame basis, here we compute it over the whole frames of the video at once, thus in a batch-mode setting. The result confirms that the two extreme configurations ``no-stability, reality'' and ``no-stability, no-reality'' show better results, on average. These performances are obtained thanks to the effect of the reset mechanism, that allows even such unstable configurations to develop good solutions. When the reset operations are disabled, we easily incurred into numerical errors due to strong oscillations, while for example, the ``stability" cases were less affected by this phenomenon.

\begin{figure}
\centering
\includegraphics[width=0.32\textwidth]{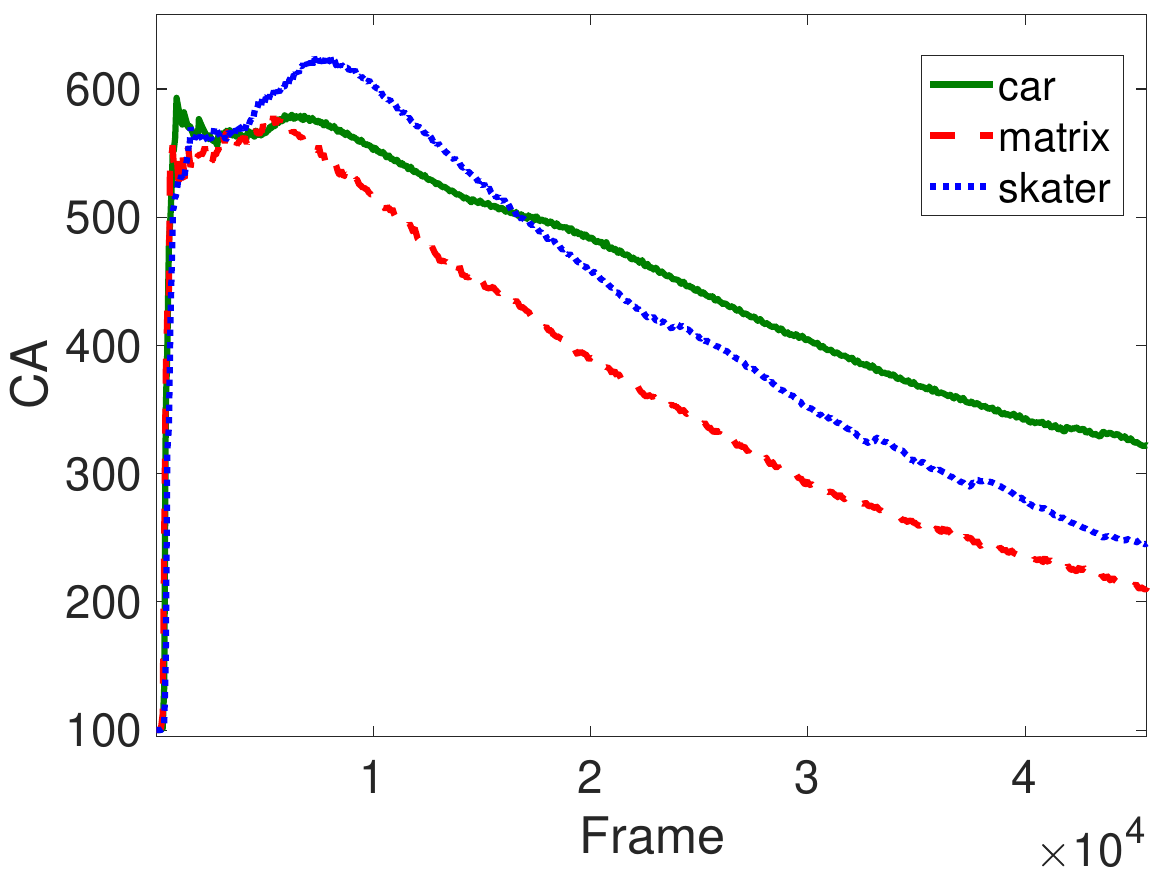}
\includegraphics[width=0.32\textwidth]{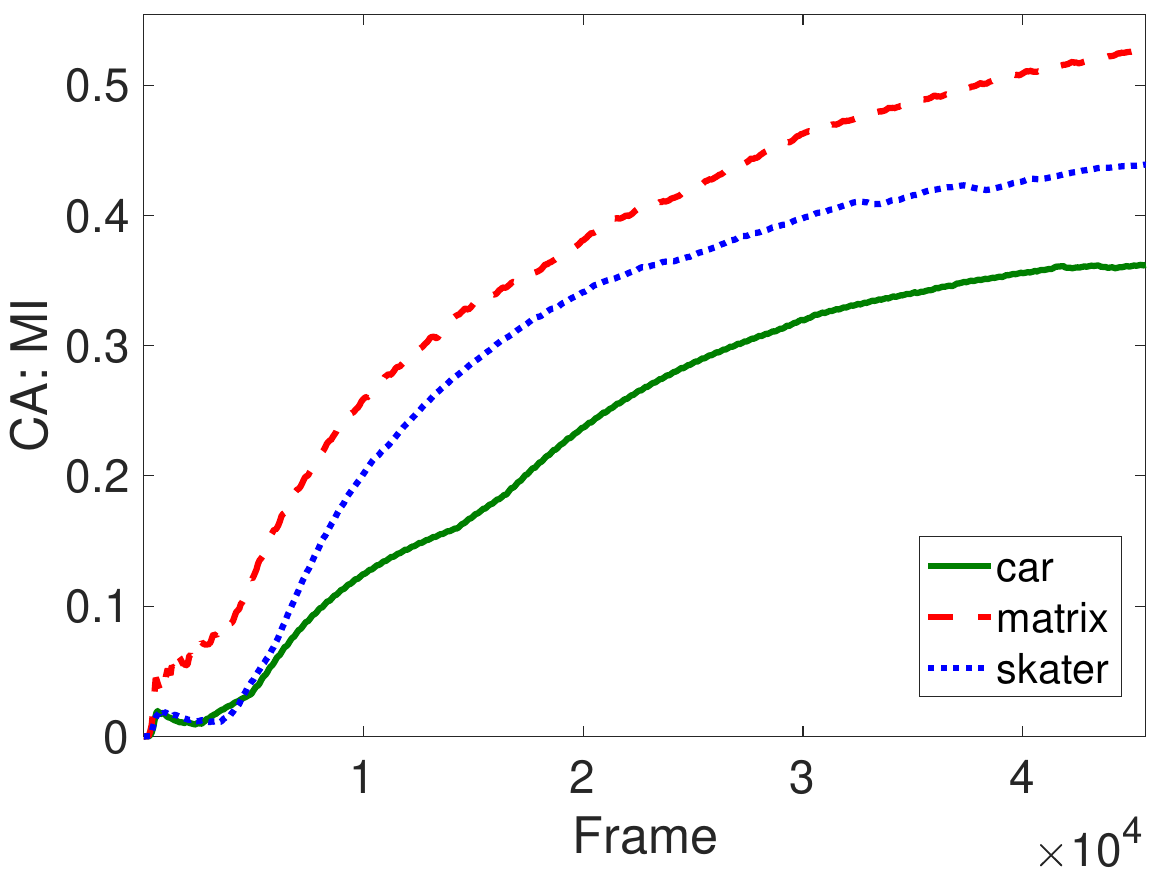}
\includegraphics[width=0.32\textwidth]{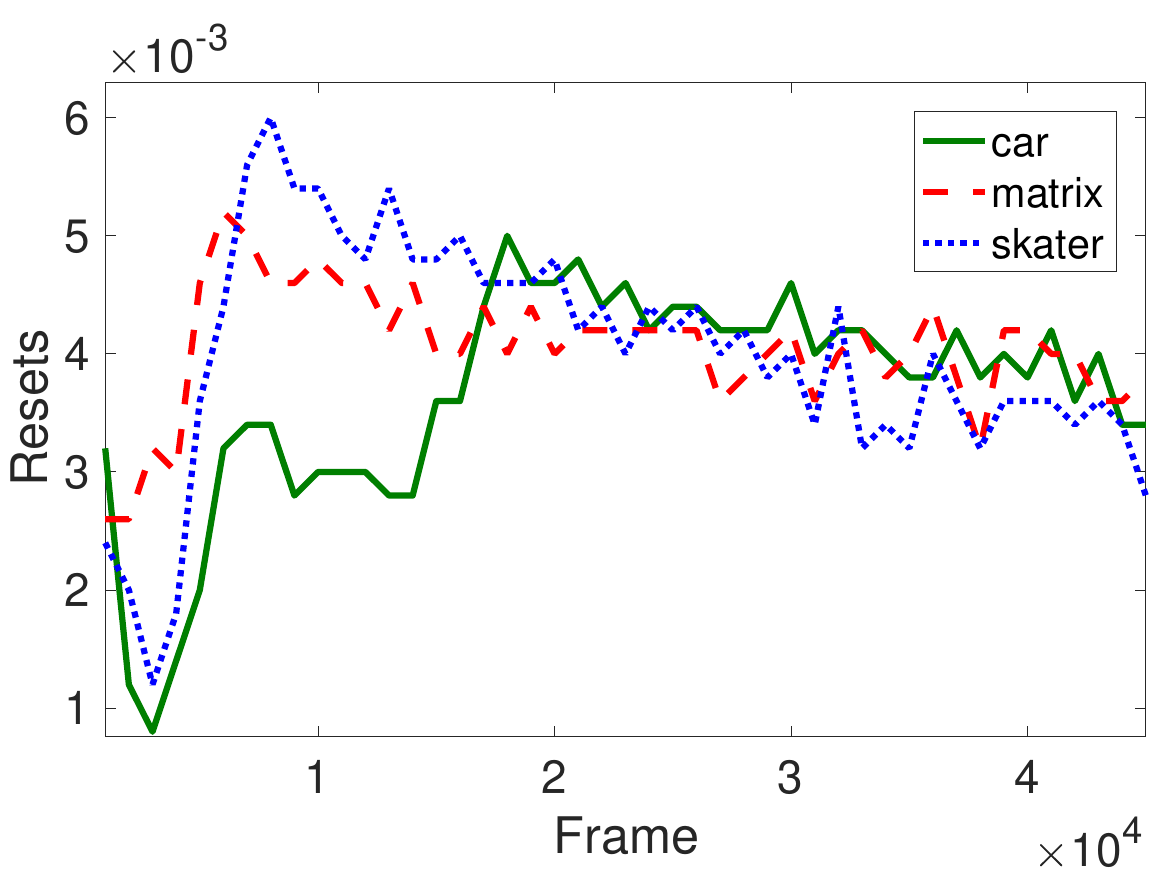}\\
\includegraphics[width=0.32\textwidth]{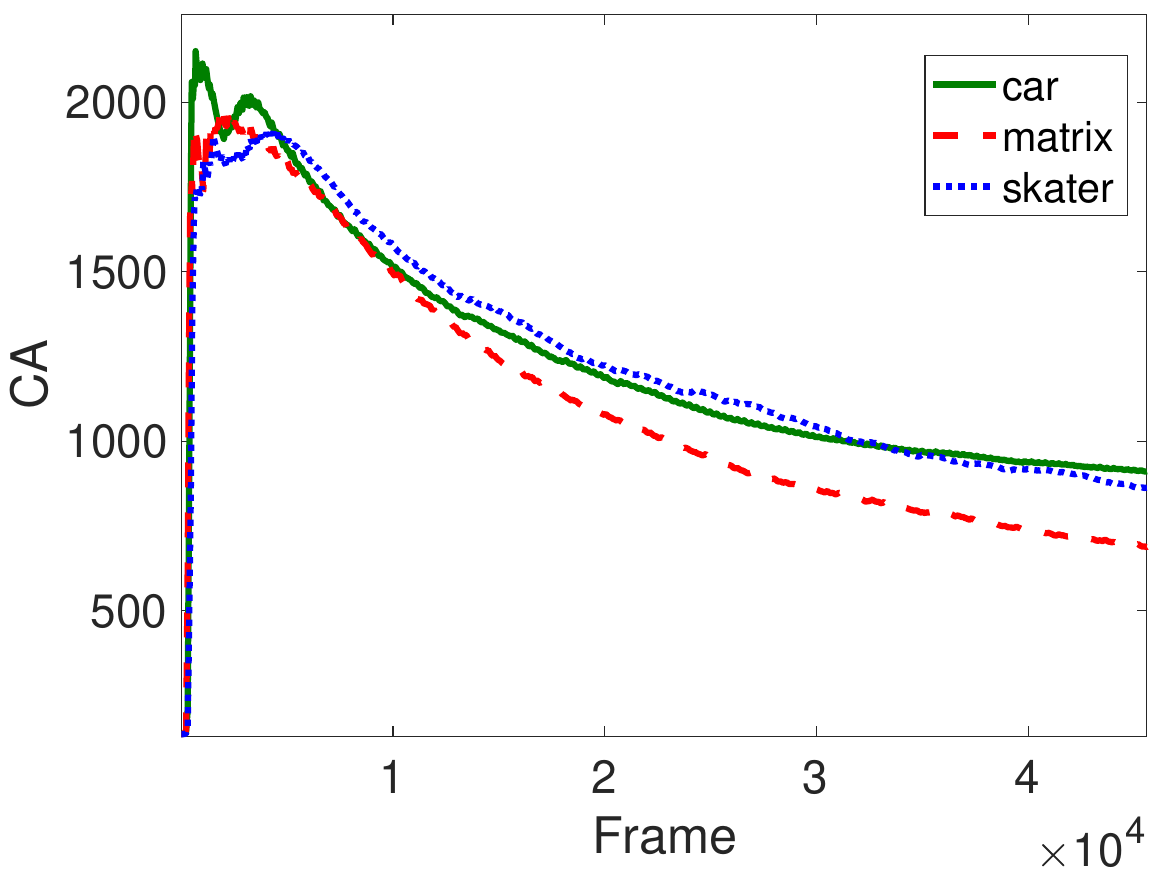}
\includegraphics[width=0.32\textwidth]{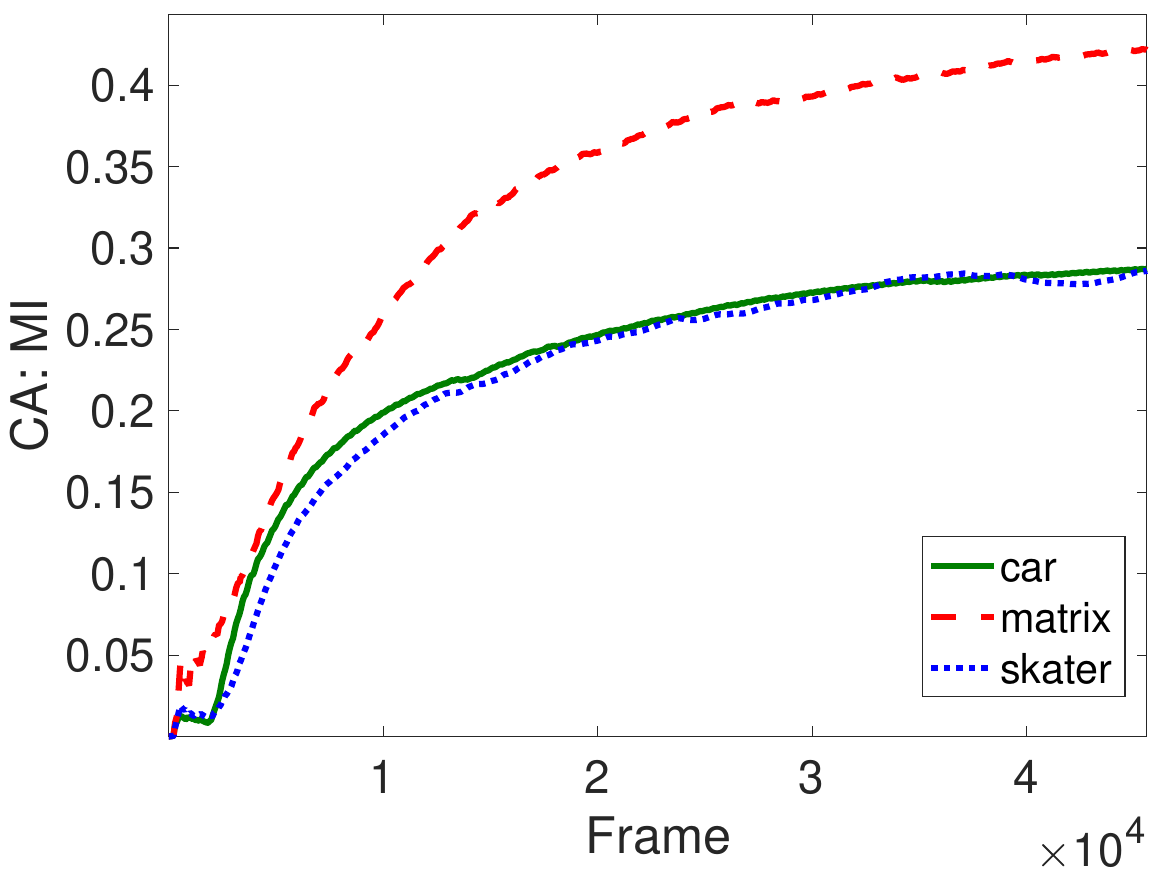}
\includegraphics[width=0.32\textwidth]{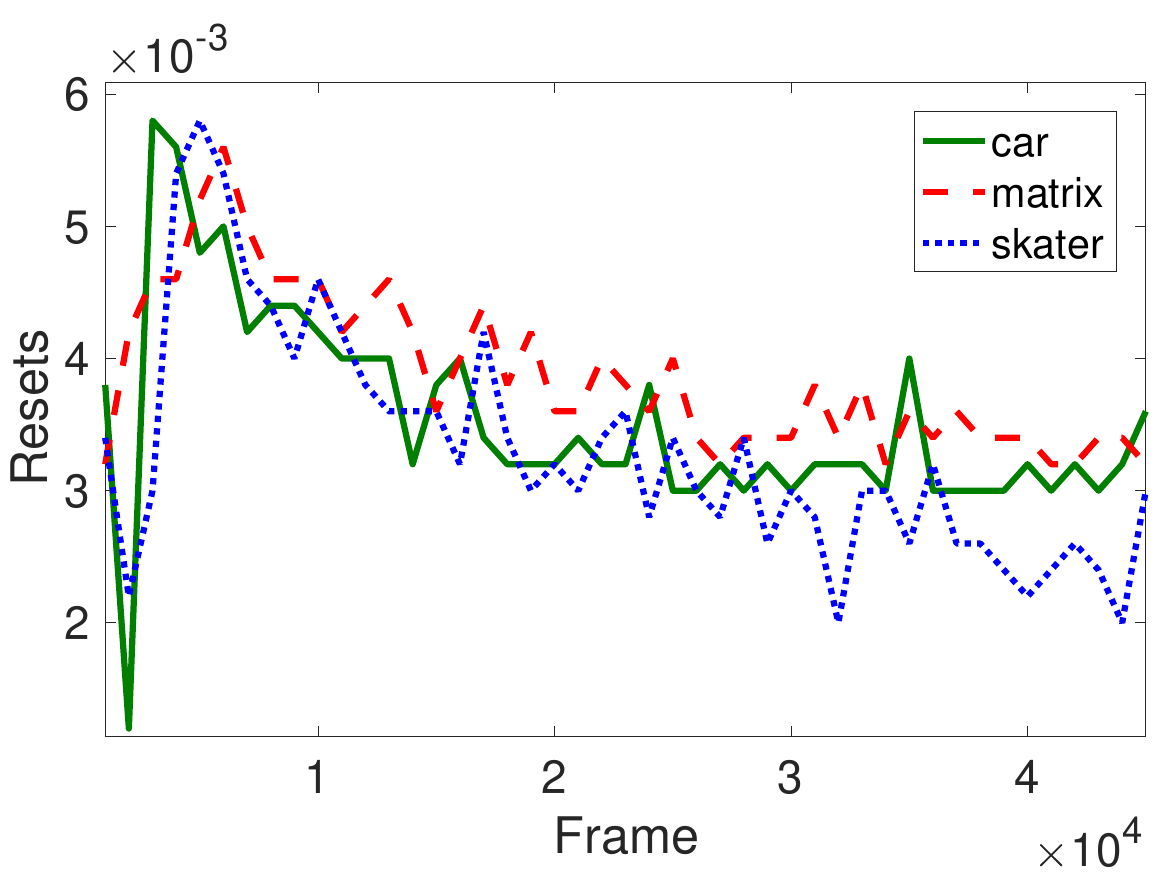}\\
\caption{Different number of features and filter sizes (1st row: $n=5, size=5\times5$; 2nd row: $n=11, size=11\times11$) in 3 videos. See Fig. \ref{smallk} for a description of the plots.}
\label{vids}
\end{figure}

We also compared the dynamics of the system on multiple video clips and using different filter sizes ($5\times5$ and $11\times11$) and number of features ($n=5$ and $n=11$) in Fig~\ref{vids}. We selected the ``stability, reality'' configuration of Fig.~\ref{smallk}, that fulfils the conditions of Eq.~(\ref{StabilityConditions}). Changing the video clip does not change the considerations we did so far, while increasing the filter size and number of features can lead to smaller MI index values, mostly due to the need of a better balancing the two entropy terms to cope with the larger number of features. The MI of Tab.~\ref{nffsb} (b) confirms this point. Interestingly, the best results are obtained in the longer video clip (``The Matrix'') that requires less repetitions of the video, being closer to the real online setting. 

\begin{figure}
\centering
\includegraphics[width=0.32\textwidth]{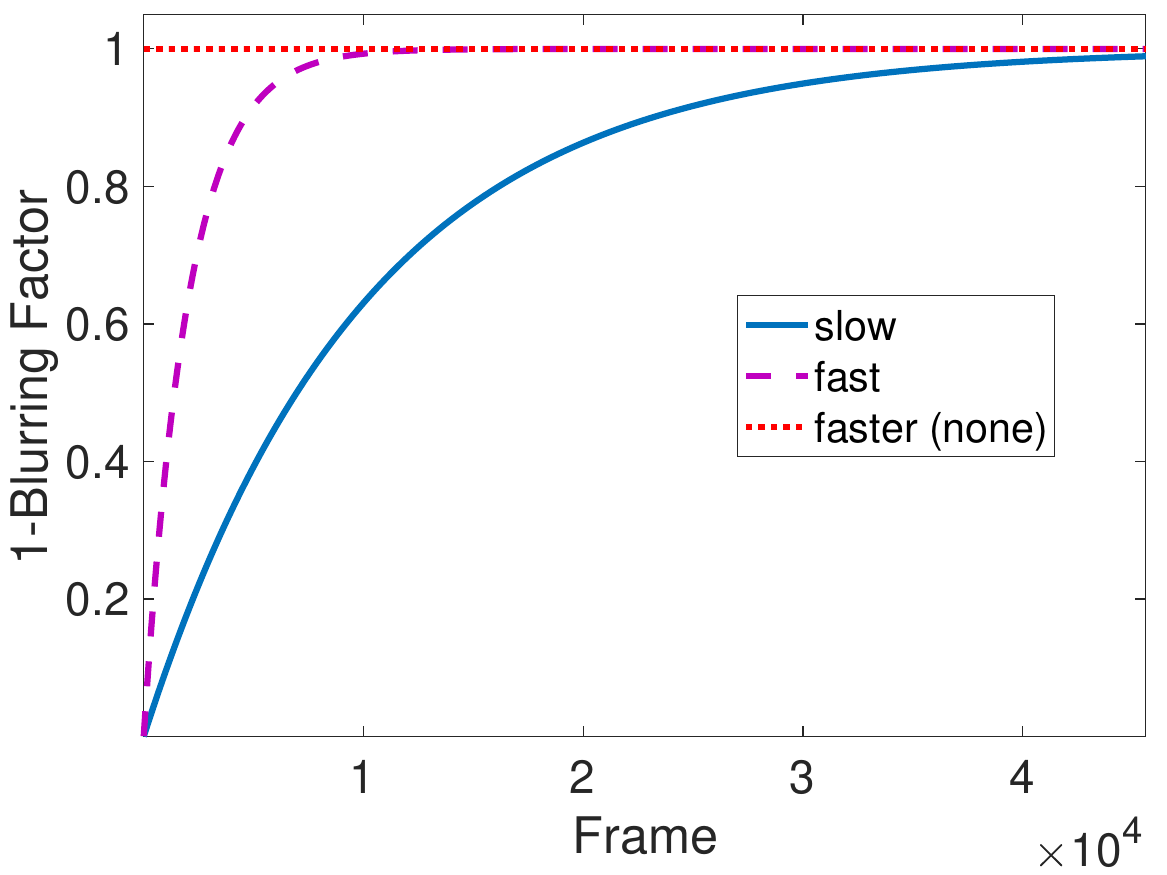}
\includegraphics[width=0.32\textwidth]{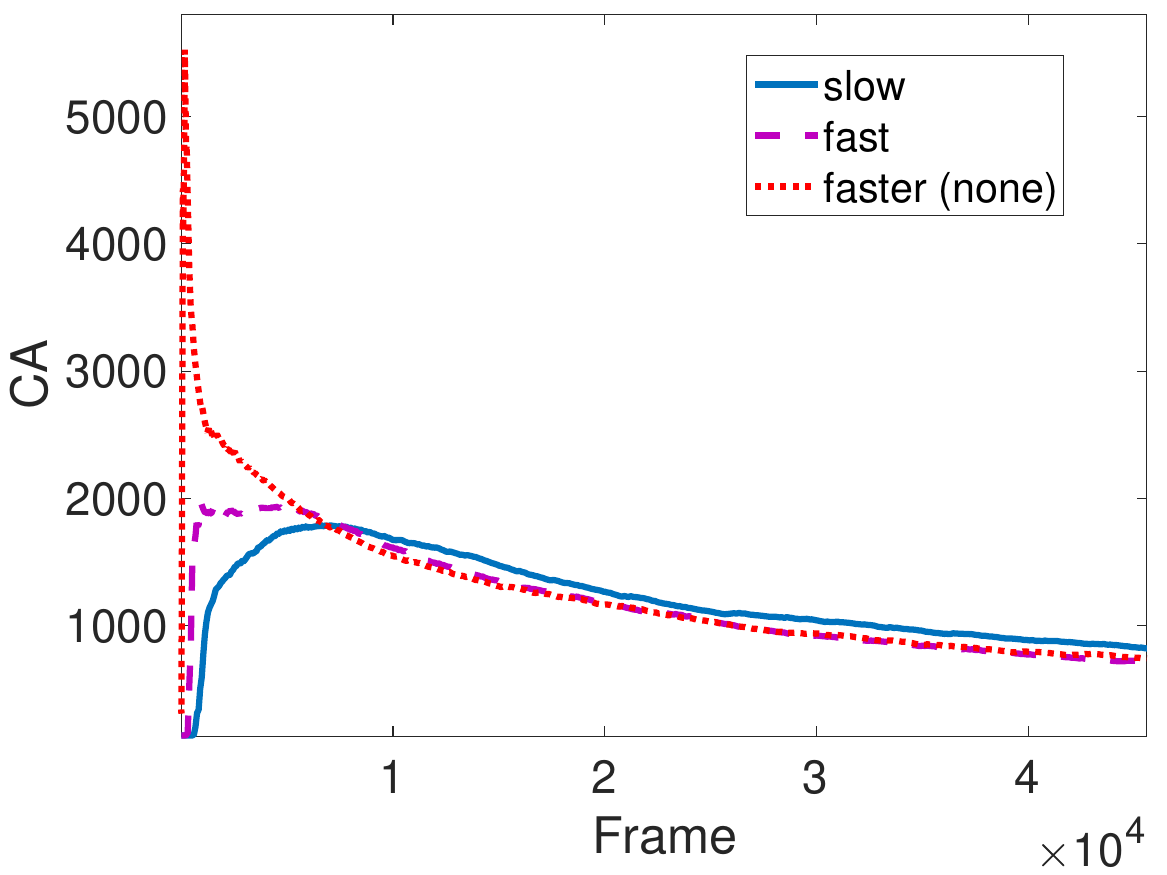}
\includegraphics[width=0.32\textwidth]{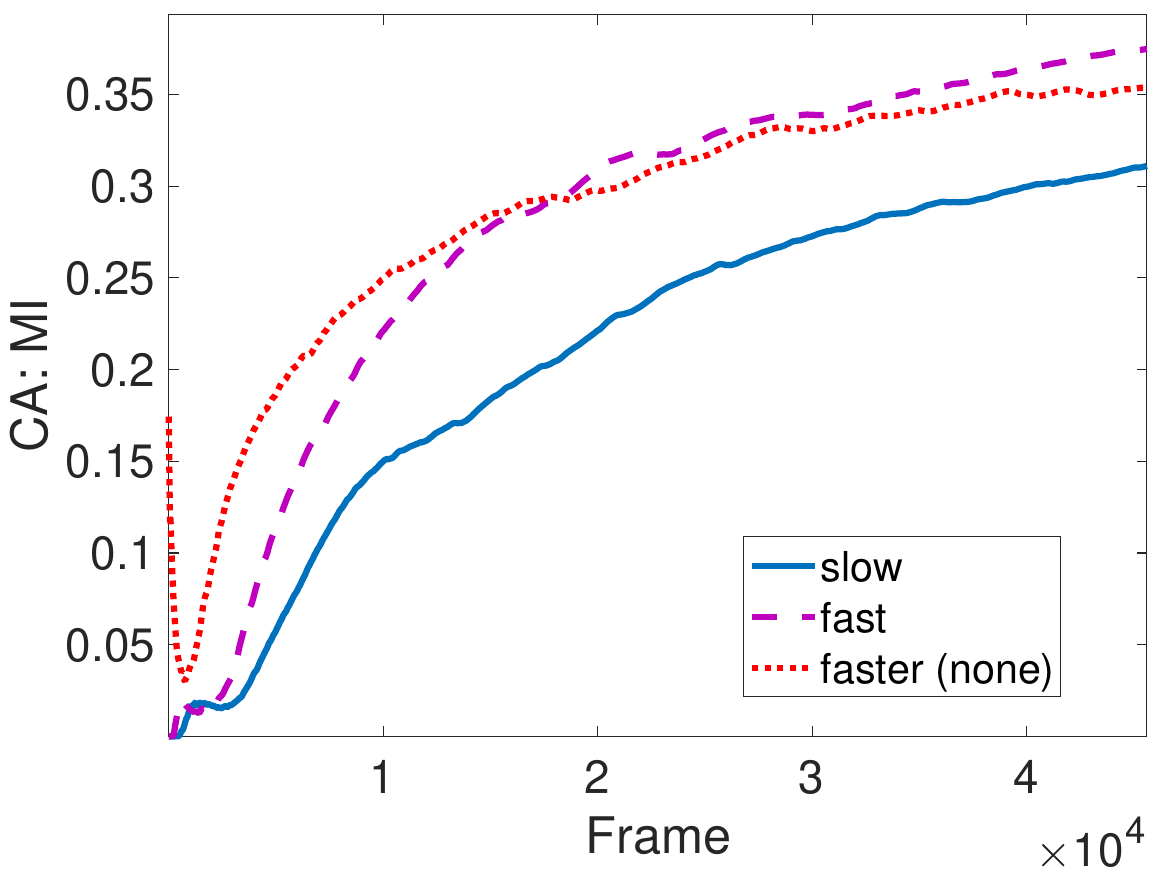}
\caption{Three different blurring plans ($n=11$ and filters of size $11\times11$).}
\label{blurs}
\end{figure}

Fig.~\ref{blurs} and Tab. \ref{nffsb} (c) show the results we obtain when using different blurring plans (``skater'' clip), that is, different values of $\eta$ that lead to the blurring factors reported in the first graph of Fig.~\ref{blurs}. These results suggest that a gradual introduction of the video signal helps the system to find better solutions than in the case in which no-plans are used, but also that a too-slow plan is not beneficial. The cognitive action has a big bump when no-plans are used, while this effect is more controlled and reduced in the case of both the slow and fast plans.


In order to study the effect of motion in multi-layer architectures (up to 3 layers), we still kept the most stable configuration (``stability, reality'', $5\times5$ filters, 5 features), and introduced the motion-related term in the cognitive action. Our multi-layer architecture is composed of a stack of computational models developed accordingly to (\ref{CognitiveActionEq}). A new layer $\ell$ is activated whenever layer $\ell-1$ has processed a large number of frames ($\approx 45k$), and the parameters of layer $\ell-1$ are not updated anymore. 
We initially considered the case in which all the layers $\ell=1,\ldots,3$ share the same value $\lambda_M$ that weighs the motion-based term. Tab. \ref{multilayer-1} shows the MI we get for different weighting schemes. Introducing motion helps in almost all the cases (for appropriate $\lambda_M$ - the smallest values of $\lambda_M$ are a good choice on average), and, as expected, a too strong enforcement of the motion-related term leads to degenerate solutions with small MI. We repeated these experiments also in a different setting. In detail, after having evaluated layer $\ell$ for all the values of $\lambda_M$, we selected the model with the largest MI and started evaluating layer $\ell+1$ on top of it. Tab. \ref{multilayer-2} reports the outcome of this experience. We clearly see that motion plays an important role in increasing the average MI. In the case of ``car'', we also obtained two (uncommon) positive results when strongly weighing $\lambda_M$. They are due to very frequent reset operations, that avoided the system to alter the filters when the motion-based term was leading to very large derivatives. This is an interesting behaviour that, however, was not common in the other cases we reported. 

\begin{table}
  \caption{MI on (a) the ``skater'' video, given the models of Fig. \ref{smallk} ($S$=stability, $R$=reality, $\bar X$=not X); (b) different videos, number of features, filter sizes (SR); (c) different blurring plans (SR).}
  \label{nffsb}
  \centering
  \begin{tabular}{rcrccrc}
    \toprule
    \multicolumn{2}{c}{(a)}&\multicolumn{3}{c}{(b)}&\multicolumn{2}{c}{(c)}\\
    \noalign{\smallskip}
    Config& \scriptsize(Skater)&Video& \scriptsize ($n=5$, $5\times 5$)& \scriptsize($n=11$, $11\times 11$)&
    Blurring& \scriptsize($n=10$, $5\times 5$)\\
    \cmidrule(r){1-2}\cmidrule(r){3-5}\cmidrule(r){6-7}
    $\bar S\bar R$ & $0.54\pm0.07$ & Car    &$0.38\pm 0.03$& $0.272\pm 0.003$ &Slow           & $0.35\pm 0.08$\\
    $\bar S R$ & $0.54\pm 0.08$ & Matrix &$0.60\pm 0.03$& $0.45\pm 0.02$   &Fast           & $0.39\pm 0.05$\\
    $S \bar R$ & $0.44\pm 0.11$ & Skater &$0.45\pm 0.13$& $0.35\pm 0.05$   &None & $0.34\pm 0.08$\\
    $SR$ & $0.45\pm 0.13$ \\
    \bottomrule
  \end{tabular}
\end{table}


\begin{table}
  \caption{MI in different videos, up to 3 layers ($\ell=1,2,3$), and for multiple weighting factors $\lambda_M$ of the motion-based term. All layers share the same $\lambda_M$.}
  \label{multilayer-1}
  \centering
  \newcolumntype{C}{>{\normalsize}c}
  \begin{tabu}{p{1mm}cccccccc}
    \toprule
    \rowfont{\small}
    &$ $& $\lambda_M=0$ &$10^{-8}$&$10^{-6}$
    &$10^{-4}$&$10^{-2}$&$1$&$10^2$\\
    \midrule\multirow{3}{*}{\rotatebox[origin=c]{90}{~Skater}}
    &$\ell=1$& $.61{\tiny \pm  .11}$ & $.54\pm .11$ &$.52\pm .07$&$.53\pm .08$
              &$\mathbf{.69}\pm .07$&$.53\pm 0$&$.01\pm 0$\\
    &$\ell=2$&$.53\pm .12$&$\mathbf{.62}\pm .15$&$.60\pm .11$&$.43\pm .06$&$.48\pm .06$
              &$.1 \pm .1$& $.03\pm .01$\\
    &$\ell=3$& $.56\pm .17$&$.58\pm .20$&$\mathbf{.62}\pm.10$& $.18\pm .16$
    &$.16\pm .17$& $.04\pm .02$&$.03\pm .02$\\
    \noalign{\smallskip}
    \multirow{3}{*}{\rotatebox[origin=c]{90}{~Car}}
    &$\ell=1$& $.49\pm.05$&$.44\pm.02$&$.46\pm.04$&$.47\pm.04$&$\mathbf{.66}\pm.10$&$.60\pm.02$
            &$.01\pm0$\\
    &$\ell=2$&$.25\pm .26$&$.54\pm.10$&$\mathbf{.65}\pm.08$&$.46\pm.03$&$.63\pm.11$&$.18\pm.32$
            &$.03\pm.01$\\
    &$\ell=3$&$.26\pm.34$&$.45\pm.22$&$\mathbf{.51}\pm.11$&$.38\pm.20$&$.24\pm.20$&$.09\pm.12$
    &$.04\pm.02$ \\
    \noalign{\smallskip}
    \multirow{3}{*}{\rotatebox[origin=c]{90}{~Matrix}}
    &$\ell=1$&$.66\pm.01$&$.66\pm.02$&$\mathbf{.67}\pm.01$&$.63\pm.05$&$.59\pm.03$&$.44\pm0$
             &$.23\pm.02$\\
    &$\ell=2$&$.55\pm.13$&$.56\pm.14$&$.43\pm0$&$.45\pm.04$&$\mathbf{.62}\pm.02$&$.35\pm.19$
             &$.13\pm.08$\\
    &$\ell=3$&$\mathbf{.64}\pm.03$&$.54\pm.11$&$.35\pm.07$&$.40\pm.01$&$.21\pm.07$&$.06\pm.03$
             &$.04\pm.02$\\
    \bottomrule
  \end{tabu}
\end{table}

\begin{table}
  \caption{Same structure of Tab. \ref{multilayer-1}. Here the model with the best $\lambda_M$ is selected and used as basis to activate a new layer (layer $\ell=1$ is the same as Tab. \ref{multilayer-1}).}
  \label{multilayer-2}
  \centering
  \newcolumntype{C}{>{\normalsize}c}
  \begin{tabu}{p{1mm}cccccccc}
    \toprule
    \rowfont{\small}
    &$ $& $\lambda_M=0$ &$10^{-8}$&$10^{-6}$
    &$10^{-4}$&$10^{-2}$&$1$&$10^2$\\
    \midrule\multirow{6}{*}{\rotatebox[origin=c]{90}{~Matrix Car Skater}}
    &$\ell=2$&$.38\pm .34$&$\mathbf{.53}\pm .12$&$.50\pm .1$&$.47\pm .1$&$.41\pm .02$
              &$.33 \pm .17$& $.21\pm .2$\\
    &$\ell=3$& $.55\pm .12$&$\mathbf{.62}\pm .11$&$.55\pm.13$& $.42\pm .01$
    &$.36\pm .09$& $.2\pm .18$&$.39\pm .22$\\
    \noalign{\smallskip}
    &$\ell=2$& $.48\pm.1$&$.59\pm.17$&$.59\pm.18$&$.55\pm.12$&$.41\pm.01$&$.01\pm0$
            &$\mathbf{.64}\pm.01$\\
    &$\ell=3$&$.67\pm.01$&$.60\pm.12$&$\mathbf{.73}\pm.09$&$.36\pm.05$&$.33\pm.11$&$.27\pm.14$
    &$\mathbf{.73}\pm.01$ \\
    \noalign{\smallskip}
    &$\ell=2$&$.55\pm.13$&$.56\pm.14$&$.43\pm0$&$.45\pm.04$&$\mathbf{.62}\pm.02$&$.35\pm.19$
             &$.13\pm.08$\\
    &$\ell=3$&$.55\pm.12$&$.53\pm.12$&$\mathbf{.82}\pm.14$&$.35\pm.05$&$.35\pm.31$&$.02\pm.01$
             &$.01\pm0$\\
    \bottomrule
  \end{tabu}
\end{table}

%% file: Conclusions.tex
%
%
In this paper we have introduced a new approach to learning visual features according
to the principle of least cognitive action. 
The experiments indicate the remarkable difference coming from the incorporation of motion
invariance, with respect to the features only driven by information-based principles, which also
results in the improvement of the mutual information from the video to the features.

%
%
The theory is coherent with the different role of the ventral stream and dorsal stream~\cite{GoodaleMilner92}
that has been observed in humans and other primates. 
The enforcement of motion invariance is clearly conceived for extracting 
features that are useful for object recognition to assolve the ``what'' task (ventral stream),
whereas ``dorsal neurons'', that are involved for where/how environmental interactions are expected not to use motion invariance.
The model behind the learning of the filters indicates the need to access to velocity estimation,
which is consistent with neuroanatomical evidence.

Although the experimental results reported in the paper assume a uniform probability distribution
in the spatiotemporal domain, the given formulation in the framework of the principle of least cognitive action 
suggests that the optimization must take place in areas of high saliency. In this case, the reformulation of the
Euler-Lagrange equations given in this paper leads to identify the crucial role of eye movements
in animals with foveal eyes.

%% file: main_nips.bbl
\begin{thebibliography}{10}

\bibitem{Marr82}
D.~Marr.
\newblock {\em Vision}.
\newblock Freeman, San Francisco, 1982.
\newblock Partially reprinted in \cite{Anderson88}.

\bibitem{imagenet_cvpr09}
J.~Deng, W.~Dong, R.~Socher, L.-J. Li, K.~Li, and L.~Fei-Fei.
\newblock {ImageNet: A Large-Scale Hierarchical Image Database}.
\newblock In {\em CVPR09}, 2009.

\bibitem{Poggio:2016:VCD}
Tomaso~A. Poggio and Fabio Anselmi.
\newblock {\em Visual Cortex and Deep Networks: Learning Invariant
  Representations}.
\newblock The MIT Press, 1st edition, 2016.

\bibitem{LoweCV2004}
D.~Lowe.
\newblock Distinctive image features from scale-invariant keypoints.
\newblock {\em International Journal of Computer Vision}, 60(2):91--110, 2004.

\bibitem{DBLP:journals/cviu/GoriLMM16}
Marco Gori, Marco Lippi, Marco Maggini, and Stefano Melacci.
\newblock Semantic video labeling by developmental visual agents.
\newblock {\em Computer Vision and Image Understanding}, 146:9--26, 2016.

\bibitem{HornAI1981}
B.~K.P. Horn and B.G. Schunck.
\newblock Determining optical flow.
\newblock {\em Artificial Intelligence}, 17(1-3):185--203, 1981.

\bibitem{Baker:2011}
Simon Baker, Daniel Scharstein, J.~P. Lewis, Stefan Roth, Michael~J. Black, and
  Richard Szeliski.
\newblock A database and evaluation methodology for optical flow.
\newblock {\em Int. J. Comput. Vision}, 92(1):1--31, March 2011.

\bibitem{DBLP:journals/tcs/BettiG16}
Alessandro Betti and Marco Gori.
\newblock The principle of least cognitive action.
\newblock {\em Theor. Comput. Sci.}, 633:83--99, 2016.

\bibitem{gori2012information}
Marco Gori, Stefano Melacci, Marco Lippi, and Marco Maggini.
\newblock Information theoretic learning for pixel-based visual agents.
\newblock In {\em European Conference on Computer Vision}, pages 864--875.
  Springer, 2012.

\bibitem{DBLP:journals/tnn/MelacciG12}
Stefano Melacci and Marco Gori.
\newblock Unsupervised learning by minimal entropy encoding.
\newblock {\em {IEEE} Trans. Neural Netw. Learning Syst.}, 23(12):1849--1861,
  2012.

\bibitem{marszalek09}
Marcin Marsza{\l}ek, Ivan Laptev, and Cordelia Schmid.
\newblock Actions in context.
\newblock In {\em IEEE Conference on Computer Vision \& Pattern Recognition},
  2009.

\bibitem{GoodaleMilner92}
Melvyn~A. Goodale and A.~David. Milner.
\newblock Separate visual pathways for perception and action.
\newblock {\em Trends in Neurosciences}, 15(1):20--25, 1992.

\end{thebibliography}
